\documentclass{article} 
\usepackage[preprint]{colm2026_conference}

\usepackage{amsmath,amssymb,amsthm}
\usepackage{graphicx}
\usepackage{booktabs}
\usepackage{microtype}
\usepackage{hyperref}
\usepackage{url}
\usepackage{xcolor}
\usepackage{enumitem}
\usepackage{caption}
\usepackage{subcaption}
\usepackage{float}
\usepackage{lineno}

\definecolor{darkblue}{rgb}{0, 0, 0.5}
\hypersetup{colorlinks=true, citecolor=darkblue, linkcolor=darkblue, urlcolor=darkblue}

\title{MetaSAEs: Joint Training with a Decomposability Penalty\\
Produces More Atomic Sparse Autoencoder Latents}

\author{
  Matthew Levinson\thanks{This work was supported by a scholarship and
  computational resources from Simplex AI Safety. The author thanks Paul
  Riechers and Adam Shai for their support.} \\
  Independent Researcher \\
  \texttt{good.epic@gmail.com}
}

\begin{document}

\ifcolmsubmission
\linenumbers
\fi

\maketitle

\begin{abstract}
Sparse autoencoders (SAEs) are increasingly used for safety-relevant applications
including alignment detection and model steering.
These use cases require SAE latents to be as atomic as possible. Each latent
should represent a single
coherent concept drawn from a single underlying representational subspace.
In practice, SAE latents blend representational subspaces together. A single
feature can activate across semantically distinct contexts that share no true
common representation, muddying an already complex picture of model computation.
We introduce a joint training objective that directly penalizes this subspace
blending.
A small \emph{meta SAE} is trained alongside the primary SAE to sparsely
reconstruct the primary SAE's decoder columns; the primary SAE is penalized
whenever its decoder directions are easy to reconstruct from the meta
dictionary. This occurs whenever latent directions lie in a subspace spanned by other
primary directions.
This creates gradient pressure toward more mutually independent decoder directions
that resist sparse meta-compression.

On GPT-2 large (layer 20), the selected configuration reduces mean $|\varphi|$ by
7.5\% relative to an identical solo SAE trained on the same data.
Automated interpretability (fuzzing) scores improve by 7.6\%,
providing external validation of the atomicity gain independent of the
training and co-occurrence metrics.
Reconstruction overhead is modest. The joint SAE increases CE loss by 3.1\%
above the baseline language model. This compares to 2.5\% for the solo SAE alone,
an additional overhead of 0.6 percentage points.
Results on Gemma 2 9B are directional. On not-fully-converged SAEs, the same
parameterization yields the best results, a $+8.6\%$ $\Delta$Fuzz.
Though directional, this is an encouraging sign that the method transfers to a larger model.
Qualitative analysis confirms that features firing on polysemantic tokens are split
into semantically distinct sub-features, each specializing in a distinct
representational subspace.
\end{abstract}

\section{Introduction}
\label{sec:intro}

Mechanistic interpretability aims to understand the computations of neural networks
in terms of human-interpretable representations and computations.
Sparse autoencoders (SAEs) have become a central tool for this effort, learning
overcomplete dictionaries of features that decompose dense model activations into
sparse, hopefully monosemantic latents~\citep{bricken2023monosemanticity,
cunningham2023sparse, templeton2024scaling}.
Beyond basic interpretability research, SAE features are increasingly used in
practical applications: detecting alignment risk,
and steering model behavior by intervening in the latent space.
These downstream uses place a premium on \emph{atomic} latents, features that
each represent one semantic feature rather than a blend of several.

\paragraph{The non-atomicity problem.}
Neural network representations are distributed across many overlapping
subspaces~\citep{elhage2022superposition}.
An ideal SAE latent would specialize to a single such subspace, making the
decomposition clean and interpretable.
In practice, SAE training does not guarantee this. A single latent can blend
representations from distinct semantic subspaces, activating on multiple unrelated
contexts that share no true common underlying representation.
This \emph{subspace blending} muddies an already complex picture of model
computation.
\citet{bussmann2024metasaes} showed this concretely. By training a secondary
(``meta'') SAE to reconstruct the decoder columns of a primary SAE, they
demonstrated that primary latents routinely decompose into combinations of
meta-latents.
A single ``Einstein'' latent, for example, decomposed into ``scientist'',
``German'', and ``famous person'' meta-latents. The primary
SAE had merged conceptually distinct representational directions into one feature.
In this particular case, Einstein may still be a good use of a dictionary element.
This demonstrates the point that an SAE may blend features together that confound
important use cases. For example, a feature conflating two concepts that is activated
when steering a model will encourage the desired behavior while also propagating
effects in other, semantically distinct contexts.

\paragraph{Our approach.}
Rather than diagnosing non-atomicity post-hoc, can we reduce it during training?
We introduce a joint training objective in which the meta SAE is trained
\emph{simultaneously} with the primary SAE.
The meta SAE continuously tries to compactly reconstruct the primary decoder
columns. The primary SAE is penalized when its decoder vectors lie in low-dimensional
subspaces already captured by the meta dictionary, making them easy to reconstruct.
This creates a gradient pressure that pushes primary features into regions of
activation space that resist sparse meta-reconstruction, driving the primary dictionary
toward greater mutual independence.

\paragraph{Contributions.}
\begin{itemize}[topsep=2pt, itemsep=1pt]
  \item A joint training objective combining primary and meta SAE training
    with a decomposability penalty (\S\ref{sec:method}).
  \item Evidence that joint training increases feature atomicity, measured
    via pairwise co-occurrence rates ($|\varphi|$) and automated interpretability
    scoring (\S\ref{sec:results}), with careful consideration of alternative
    explanations (\S\ref{sec:alternative}).
  \item Qualitative case studies demonstrating semantically
    distinct feature specialization (\S\ref{sec:qualitative}).
  \item Positive directional results on Gemma 2 9B (\S\ref{sec:gemma}) with the
  same winning parameterization as GPT-2 large.
\end{itemize}

\section{Background}
\label{sec:background}

\paragraph{Sparse autoencoders.}
An SAE learns an encoder $f_\text{enc}:\mathbb{R}^d\!\to\!\mathbb{R}^n$ and
decoder $f_\text{dec}:\mathbb{R}^n\!\to\!\mathbb{R}^d$ with $n \gg d$ and a
sparse latent code.
We use BatchTopK SAEs~\citep{gao2024scaling}. Given a batch of activations,
the top-$k$ pre-activations across \emph{all} latents in the batch are retained,
enforcing an exact batch-level L0 of $k$.
The $i$-th decoder column $W_\text{dec}[i] \in \mathbb{R}^d$ is the ``feature
direction'' for latent $i$.
SAEs have been applied to GPT-2~\citep{cunningham2023sparse}, Claude models~\citep{templeton2024scaling},
and Gemma~\citep{lieberum2024gemma}, and have been used to identify safety-relevant internal features,
including ones associated with deception, bias, sycophancy, and dangerous content~\citep{templeton2024scaling}.

\paragraph{Feature non-atomicity and superposition.}
\citet{elhage2022superposition} argued that neural networks represent more features
than they have dimensions by encoding them as directions in an overcomplete basis.
This superposition hypothesis implies that a single SAE latent may be pulled toward
directions that blend multiple underlying features, particularly when the training
objective (reconstruction fidelity plus sparsity) does not directly penalize such
blending.
\citet{bussmann2024metasaes} introduced meta SAEs: a secondary SAE trained on
the decoder columns of a primary SAE, treating each column as an input vector.
A small meta dictionary ($m \ll n$) is forced to find shared structure across
primary features. Low meta-reconstruction error for primary feature $i$ indicates
that $W_\text{dec}[i]$ lies near the subspace spanned by the current meta dictionary.
In their setup the meta SAE is trained sequentially on a frozen primary, so no
signal flows back to improve it.

\paragraph{Orthogonality penalties.}
\citet{korznikov2025ortsae} (OrtSAE) pursue a related goal via a direct
orthogonality penalty on primary SAE decoder columns, penalizing high pairwise
cosine similarity within chunks of the decoder matrix.
OrtSAE and our method share the motivation of pushing feature directions apart,
but differ in approach. OrtSAE enforces pairwise geometric separation
uniformly across all decoder column pairs, whereas our decomposability penalty
targets \emph{sparse reconstructability}.
This distinction means our penalty is more sensitive to the semantic structure of the
feature space, rather than treating all pairs symmetrically.

\paragraph{Automated interpretability and fuzzing.}
\citet{bills2023language} introduced the use of language models to generate
natural-language descriptions of neural network features and evaluate their
quality.
\citet{paulo2024autointerp} extended this to the \emph{fuzzing}
paradigm. Given a candidate description of a feature, a judge model must identify
which of two activation examples is real versus randomly modified (highest-activating
tokens replaced by random draws from the same marginal distribution).
Scores range from $-1$ (consistently wrong) to $+1$ (always correct), with $0$
at chance.
Fuzzing is particularly sensitive to atomicity. A blended feature activating on
two unrelated concepts cannot be coherently described in a way that reliably
predicts which examples are real. So blended features tend to score poorly
compared to single-concept features.

\paragraph{The phi coefficient as a co-occurrence metric.}
For two binary firing indicators (latent $i$ fires / does not fire, latent $j$
fires / does not fire), the phi ($\varphi$) coefficient measures co-occurrence
above chance:
\begin{equation}
  \varphi(i,j) = \frac{N_{11}\,n - N_{1\cdot i}\,N_{1\cdot j}}
    {\sqrt{N_{1\cdot i}(n-N_{1\cdot i})\,N_{1\cdot j}(n-N_{1\cdot j})}},
  \label{eq:phi}
\end{equation}
where $N_{11}$ is the number of tokens on which both latents fire,
$N_{1\cdot i},N_{1\cdot j}$ are marginal firing counts, and $n$ is total tokens.
We use mean $|\varphi|$ as our primary co-occurrence metric throughout.
Crucially, $\varphi$ is normalized by the marginal firing rates: changes in how
often individual features fire do not by themselves change $\varphi$.

\paragraph{Why mean $|\varphi|$ as a grid selection criterion.}
One might ask whether reducing mean $|\varphi|$ is a meaningful goal:
splitting a polysemantic solo feature into two joint features could increase
or decrease the total co-occurrence depending on the specifics of the split.
We argue that the dominant type of co-occurrence in a large overcomplete
dictionary is \emph{subtle subspace blending}. Many feature pairs have small but
non-zero $|\varphi|$. Subspaces are represented dissimilarly but not fully independently.
If the decomposability penalty succeeds in pushing decoder directions toward
greater mutual independence, this mass of small non-zero correlations should
decrease, and the numerical dominance of these many pairs in the mean
will reliably drive mean $|\varphi|$ down even if a small number of
well-separated ``big-star'' splitting candidates move in either direction.
Fuzzing scores serve as independent external validation of this claim (\S\ref{sec:autointerp}).

\section{Method}
\label{sec:method}

\subsection{Architecture}

We train a \emph{primary SAE} on model activations and a \emph{meta SAE} on the
primary SAE's decoder columns simultaneously.
Both use BatchTopK.
The primary SAE has dictionary size $n$. The meta SAE has dictionary size
$m \ll n$. We follow \citet{bussmann2024metasaes} in using a compression
ratio of $m/n \approx 9\%$ (1,800 meta-features for a 20,480-feature primary
dictionary; their ratio was $\approx 5\%$).
With $m \ll n$, the meta SAE cannot memorize individual decoder columns. It must
find directions that are reused across the primary dictionary.

\subsection{Decomposability penalty}

Let $\hat{W}_\text{dec}[i]$ be the meta SAE's reconstruction of primary decoder
column $i$.
We define a per-feature penalty that is large when reconstruction error is small:
\begin{equation}
  p_i = \exp\!\left(-\frac{\|W_\text{dec}[i] - \hat{W}_\text{dec}[i]\|^2}{\sigma^2}\right),
  \label{eq:penalty}
\end{equation}
and augment the primary SAE's loss:
\begin{equation}
  \mathcal{L}_\text{primary} = \mathcal{L}_\text{SAE}(x) \;+\;
    \lambda_2\cdot\frac{1}{n}\sum_{i=1}^{n} p_i.
  \label{eq:loss}
\end{equation}
$\mathcal{L}_\text{SAE}$ is the standard L2 reconstruction loss plus an auxiliary
dead-feature loss.
$\lambda_2$ controls the strength of the penalty; $\sigma^2$ is a bandwidth
parameter that determines how sharply the penalty distinguishes low- from
high-error reconstructions.

Intuitively: $p_i \approx 1$ when the meta SAE reconstructs feature $i$ easily
(its decoder direction lies in the meta-feature subspace). $p_i \approx 0$ when
feature $i$ is genuinely novel relative to the meta dictionary.
Minimizing $\sum_i p_i$ pushes the primary dictionary toward features that
are more mutually independent and resist sparse meta-compression.

\subsection{Training procedure and gradient flow}

Training alternates at each macro-step:
\begin{enumerate}[topsep=2pt, itemsep=1pt]
  \item \textbf{Primary phase} ($n_p$ steps): compute the primary SAE forward
    pass and the decomposability penalty, treating the current meta SAE as a
    frozen critic; update primary SAE parameters.
  \item \textbf{Meta phase} ($n_m$ steps): use the current primary decoder matrix
    $W_\text{dec}$ as training data; update the meta SAE to minimize its
    reconstruction loss on these vectors.
\end{enumerate}
In Phase~1, $\hat{W}_\text{dec}[i]$ is computed inside \texttt{torch.no\_grad()}
so gradients flow \emph{to} $W_\text{dec}$ but not \emph{through} the meta SAE;
the resulting gradient points each decoder column away from its current
meta-reconstruction, into regions the meta dictionary has not yet learned.
As the primary decoder updates, the meta SAE adapts in Phase~2, creating an adversarial
dynamic akin to diversity regularization in ensemble methods~\citep{krogh1995neural},
but operating on the geometry of the feature dictionary.
The result is a set of primary features that are more mutually independent: each occupies
a direction not predictable from the others, reducing the subspace blending that underlies
non-atomicity.

\section{Experimental setup}
\label{sec:setup}

\paragraph{Model and layer.}
We train SAEs on residual stream activations at layer 20 (\texttt{resid\_pre})
of GPT-2 large~\citep{radford2019language} ($d = 1280$, 36 layers).
Generalization experiments use Gemma 2 9B~\citep{gemma2024} at layer 23 ($d = 3584$).

\paragraph{Dataset.}
FineWeb 10BT~\citep{fineweb2024}, streamed with context length 128.
SAE training uses 100M tokens; co-occurrence evaluation uses 35M fresh tokens.

\paragraph{SAE architecture.}
Primary SAE: $n = 20{,}480$ features, batch $k = 64$.
Meta SAE: $m = 1{,}800$ features, batch $k = 8$.
We chose BatchTopK~\citep{gao2024scaling} over JumpReLU~\citep{rajamanoharan2024jumping} via an empirical ablation comparing both
architectures across 8 configurations. BatchTopK achieved lower L2 loss in
both joint and solo conditions and was used for all subsequent experiments.

\paragraph{Hyperparameter sweep.}
We sweep $\lambda_2 \in \{0.1, 0.3, 1.0\}$ and
$\sigma^2 \in \{0.3, 1.0, 3.0\}$ (9 configurations), each trained for 100M
tokens with $n_p = 10$ primary steps and $n_m = 5$ meta steps per macro-step,
Adam with lr $= 3\times10^{-4}$.
The solo baseline is an identically configured BatchTopK SAE trained without the
meta SAE or penalty.
We select the best configuration on mean $|\varphi|$ from the 35M-token
co-occurrence scan while considering reconstruction quality. Automated interpretability scores serve as independent
validation of the selected configuration (\S\ref{sec:autointerp}).

\paragraph{Co-occurrence evaluation.}
Per-feature firing thresholds are calibrated using an equal-error-rate (EER)
procedure on 5M tokens~(Appendix~\ref{app:calibration}), giving reproducible
per-token binary firing decisions at inference time independent of batch composition.

\paragraph{Automated interpretability.}
We compute fuzzing scores~\citep{bills2023language,paulo2024autointerp} using
Qwen2.5-72B-Instruct-AWQ as the judge model.
We align joint and solo features via Hungarian matching on decoder cosine
similarity, yielding 20,456 matched pairs, and report paired $\Delta$-scores.

\section{Results}
\label{sec:results}

\subsection{Feature co-occurrence}
\label{sec:cooccurrence}

\begin{figure}[t]
  \centering
  \begin{subfigure}[b]{0.48\textwidth}
    \includegraphics[width=\textwidth]{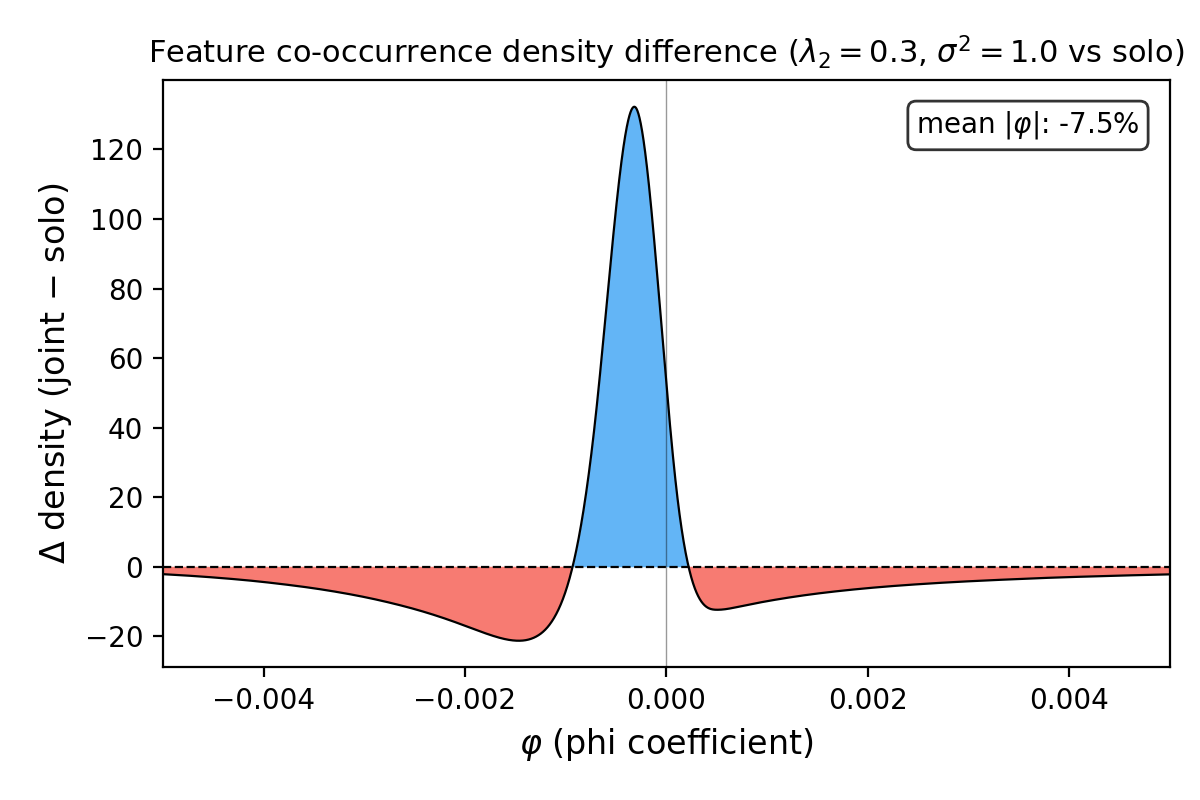}
    \caption{Histogram difference (joint $-$ solo) in $|\varphi|$ for the best
      configuration ($\lambda_2\!=\!0.3$, $\sigma^2\!=\!1.0$). Negative bars
      indicate probability mass shifted away from correlated co-activation}
    \label{fig:phi_diff}
  \end{subfigure}
  \hfill
  \begin{subfigure}[b]{0.48\textwidth}
    \includegraphics[width=\textwidth]{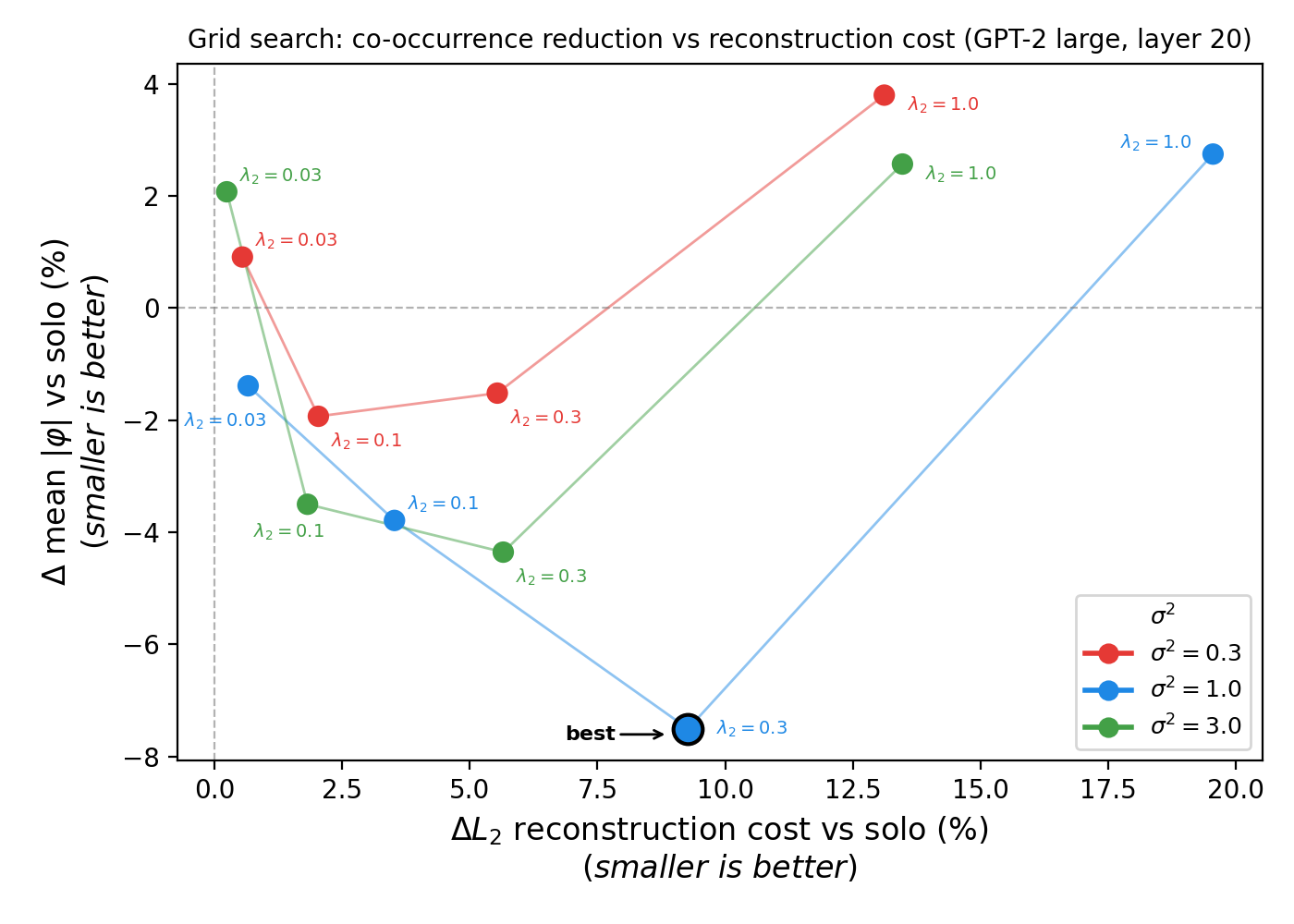}
    \caption{Tradeoff between co-occurrence reduction ($\Delta|\varphi|$) and
      reconstruction cost ($\Delta$L2) across all 12 grid configurations.
      $\lambda_2\!=\!1.0$ (red) achieves poor tradeoffs; the best Pareto point
      is $\lambda_2\!=\!0.3$, $\sigma^2\!=\!1.0$.}
    \label{fig:scatter}
  \end{subfigure}
  \caption{Co-occurrence results on GPT-2 large, layer 20.}
  \label{fig:cooccurrence}
\end{figure}

\begin{figure}[t]
\begin{minipage}[c]{0.50\textwidth}
  \centering
  \small
  \begin{tabular}{ccrrc}
  \toprule
  $\lambda_2$ & $\sigma^2$ & $\Delta\overline{|\varphi|}$ (\%) & $\Delta$L2 (\%) & $\Delta$CE (\%) \\
  \midrule
  0.1  & 0.3 & $-1.9$  & $+2.0$ & — \\
  0.1  & 1.0 & $-3.8$  & $+3.5$ & $+2.8$ \\
  0.1  & 3.0 & $-3.5$  & $+1.8$ & $+2.7$ \\
  0.3  & 0.3 & $-1.5$  & $+5.5$ & — \\
  \textbf{0.3} & \textbf{1.0} & $\mathbf{-7.5}$ & $\mathbf{+9.3}$ & $\mathbf{+3.1}$ \\
  0.3  & 3.0 & $-4.4$  & $+5.7$ & $+2.9$ \\
  1.0  & 0.3 & $+3.8$  & $+13.1$ & — \\
  1.0  & 1.0 & $+2.8$  & $+19.6$ & $+3.5$ \\
  1.0  & 3.0 & $+2.6$  & $+13.5$ & $+3.2$ \\
  \bottomrule
  \end{tabular}
  \captionof{table}{Co-occurrence and reconstruction across the hyperparameter grid, relative
  to the solo baseline ($\overline{|\varphi|}=2.27\!\times\!10^{-3}$, L2$=0.097$).
  $\Delta$CE is joint CE overhead as \% of baseline language-model CE (3.47 nats);
  solo overhead is 2.4\%. Configs without CE evaluation marked~``—''. Best row \textbf{bold}.}
  \label{tab:grid}
\end{minipage}
\hfill
\begin{minipage}[c]{0.46\textwidth}
  \centering
  \includegraphics[width=\textwidth]{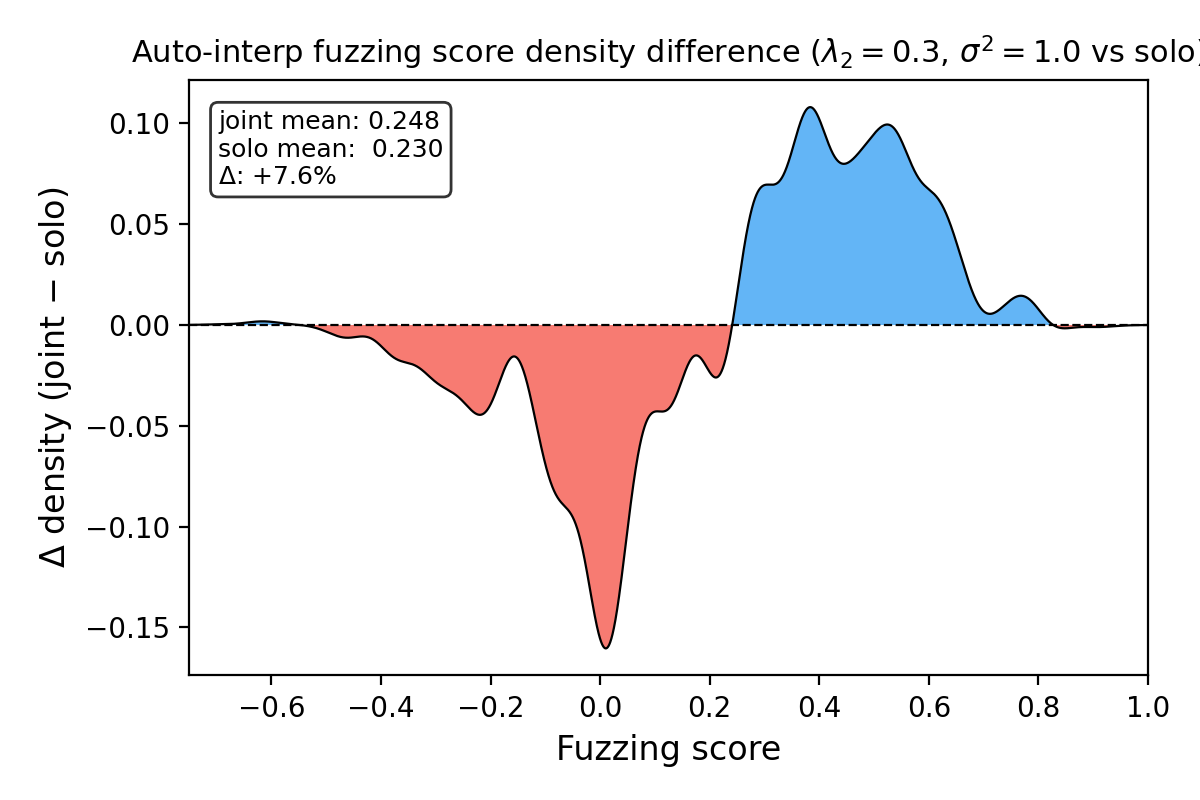}
  \captionof{figure}{Density difference (joint $-$ solo) in fuzzing score distributions.
    Blue: higher density in joint; red: higher density in solo.
    Joint training shifts mass from near-zero scores toward higher values,
    indicating more consistently interpretable features.}
  \label{fig:fuzzing}
\end{minipage}
\end{figure}

In Table~\ref{tab:grid} we report mean $|\varphi|$ and
L2 across all 12 grid configurations.
The best configuration ($\lambda_2\!=\!0.3$, $\sigma^2\!=\!1.0$) achieves a
7.5\% reduction in mean~$|\varphi|$, with $+9.3\%$ L2 increase.
The effect is highly significant. Welch's $t$-test over all $\approx$209M pairs
gives $t = {-53.1}$, $p \ll 10^{-100}$.

We verified all metrics for a $\lambda_2\!=\!0$ control (joint architecture,
no penalty), confirming no systematic effect on $|\varphi|$, L2, or fuzzing
scores relative to the solo baseline.
The effect is therefore attributable to the decomposability penalty itself,
not to the joint optimization structure.
$\lambda_2 = 1.0$ degrades reconstruction severely without
commensurate $|\varphi|$ gains.
Over-penalizing decoder reconstruction
forces directions apart even when shared geometry reflects genuine co-structure
in the data, not superficial blending.
The Pareto-optimal region, $\lambda_2 \in \{0.1, 0.3\}$, offers the best
$|\varphi|$ reduction per unit reconstruction cost
(Figure~\ref{fig:scatter}).

\paragraph{Independent baseline.}
Four independent solo SAEs yield mean~$|\varphi| \approx 2.15\times10^{-3}
\pm 10^{-5}$---natural run-to-run variation far smaller than the
$1.7\times10^{-4}$ joint-solo gap, confirming the effect is specific to joint
training rather than stochastic optimization noise.

\subsection{Automated interpretability}
\label{sec:autointerp}

Mean fuzzing score on the jointly trained primary SAE is $0.248$, compared to
$0.230$ for the solo SAE (paired $\Delta=+0.0175$, $+7.6\%$;
$t=8.12$, $p=5\times10^{-16}$); 51.6\% of pairs show higher joint scores
(Figure~\ref{fig:fuzzing}).
As argued in \S\ref{sec:background}, this constitutes external validation of
the atomicity claim independent of the co-occurrence analysis.

\subsection{Reconstruction quality}
\label{sec:reconstruction}

For the best configuration ($\lambda_2\!=\!0.3$, $\sigma^2\!=\!1.0$):
\begin{itemize}[topsep=2pt, itemsep=1pt]
  \item \textbf{CE overhead}: the solo SAE increases CE loss by 2.5\% above the
    baseline language model (absolute increase of 0.087 nats over baseline CE of 3.47).
    The joint SAE increases CE by 3.1\%, an additional overhead of 0.6 percentage
    points above the solo.
  \item \textbf{L2 loss}: $+9.3\%$ relative increase; absolute L2 increase
    $= 0.010$.
  \item \textbf{Absolute scale}: mean activation norm $\|x\|_2 = 133.3$ at this
    layer; the additional L2 cost is $0.0075\%$ of mean norm---negligible in
    absolute terms.
  \item \textbf{L0}: $64$ for both conditions, enforced by BatchTopK.
\end{itemize}

For reconstruction-sensitive applications, $\lambda_2\!=\!0.1$, $\sigma^2\!=\!3.0$
offers a better tradeoff: $-3.5\%$ $\Delta|\varphi|$ at only $+1.8\%$ $\Delta$L2.

\subsection{Qualitative case studies (GPT-2 large)}
\label{sec:qualitative}

The \emph{cross-$|\varphi|$ matrix} records $|\varphi|$ between every solo
feature and every joint feature (co-occurrence correlations across the
two SAEs rather than within one), and identifies \emph{splitting candidates}.
These are solo features that co-activate strongly with multiple semantically distinct joint
features, suggesting the solo SAE blended subspaces that the joint SAE
represents as separate directions.
We inspect splitting candidates from the $\lambda_2\!=\!0.3$,
$\sigma^2\!=\!1.0$ GPT-2 large pipeline. Full feature descriptions and
scores are in Tables~\ref{tab:qual_too} and~\ref{tab:qual_let}.

Two cases illustrate the pattern.
First, the English word \textit{too} is polysemantic. It can mean
``excessively'' (degree modifier, e.g.\ \textit{it was too hot}) or
``also / additionally'' (sentence-final, e.g.\ \textit{me too}).
The solo SAE conflates both senses in a single feature (fuzz\,$+0.727$),
blending two distinct semantic subspaces.
The joint SAE resolves them: feat.\ 4741 fires on the excess/degree-modifier
use in contexts of regret or negative outcome (fuzz\,$+0.681$,
$|\varphi|\!=\!0.710$), while feat.\ 14213 captures the sentence-final
additive use (fuzz\,$+0.174$, $|\varphi|\!=\!0.799$).

Second, the word \textit{let} serves at least three distinct pragmatic roles:
direct suggestion or imperative (``Let's go''), discourse-transition pivot
(``Let me now turn to\ldots''), and permissive allowing.
The solo SAE merges all three into one feature (fuzz\,$+0.641$),
conflating subspaces with different pragmatic and syntactic properties.
The joint SAE assigns each role to a distinct feature: feat.\ 7228 takes
the directive/invitation use ($|\varphi|\!=\!0.692$, fuzz\,$+0.646$),
feat.\ 10152 the discourse-pivot use ($|\varphi|\!=\!0.486$,
fuzz\,$+0.273$), and feat.\ 18650 the permissive-allowing use
($|\varphi|\!=\!0.435$, fuzz\,$+0.497$).
Both cases are consistent with the mutual independence hypothesis. The
decomposability penalty encourages the joint SAE to disentangle uses of the
same surface form that occupy distinct representational subspaces.

\section{Generalization to Gemma 2 9B}
\label{sec:gemma}

We apply the joint training method to Gemma 2 9B~\citep{gemma2024} (layer 23,
$n\!=\!65{,}536$ features, L0\,=\,100) sweeping
$\lambda_2 \in \{0.1, 0.3, 1.0\}$ at $\sigma^2 \in \{1.0, 3.0\}$ on 75M FineWeb tokens.
We emphasize that these are well-optimized but \emph{not fully converged} SAEs,
and the grid co-occurrence scan uses a smaller token budget (5M tokens over a
65,536$^2$ feature space), giving wide confidence intervals.
Gemma results should be interpreted as directional.

\paragraph{Config selection.}
The co-occurrence grid on Gemma is less conclusive than on GPT-2 large due to the
underpowered scan; $|\varphi|$ effects are small and mixed across configurations.
Notably, $\lambda_2\!=\!1.0$ shows the same failure mode as on GPT-2 large---
substantial $+|\varphi|$ increases alongside heavy L2 overhead---confirming this
regime is problematic regardless of model scale.
We select $\lambda_2\!=\!0.3$, $\sigma^2\!=\!1.0$ as the primary Gemma
configuration, identical to the GPT-2 large winner.
Encouragingly, this is the only Gemma configuration that shows meaningful positive
automated interpretability gains (\S\ref{sec:gemma_fuzz} below).

\paragraph{Reconstruction overhead.}
The L2 overhead pattern on Gemma is consistent with GPT-2 large and scales
with $\lambda_2$: $+8.2\%$ at $\lambda_2\!=\!0.3$, $\sigma^2\!=\!1.0$
(compared to $+9.3\%$ on GPT-2 large at the same setting).
For the selected parameters, we continued training from the grid scan
checkpoint for an additional 300M tokens. This better trained joint SAE adds
a modest additional overhead of 0.0196 CE units ($+2.5\%$)
above the solo SAE's 0.1602 CE increase ($+2.2\%$) over baseline ($7.12$ nats).

\paragraph{Automated interpretability.}
\label{sec:gemma_fuzz}
On the grid fuzz sweep ($n\!\approx\!1{,}000$ random features per configuration),
$\lambda_2\!=\!0.3$, $\sigma^2\!=\!1.0$ is the only Gemma configuration with
clearly positive $\Delta$Fuzz at $+8.6\%$, directionally consistent with
the GPT-2 large result.
The fact that the best Gemma configuration matches the GPT-2 large winner
without any Gemma-specific tuning is encouraging evidence that
$\lambda_2\!\approx\!0.3$, $\sigma^2\!\approx\!1.0$ may be a robust default.

\paragraph{Qualitative examples.}
We identify splitting candidates from the Gemma 2 9B cross-$|\varphi|$ scan
using the pipeline described in~\ref{sec:qualitative}. See details in
Tables~\ref{tab:qual_cookies} and~\ref{tab:qual_flash} in the appendix.

The word \textit{cookies} illustrates a clean two-way sense split.
The solo SAE conflates the food sense (\textit{baked cookies}) and the
web-technology sense (\textit{browser cookies, data tracking}) in a single
feature (fuzz\,$+0.493$).
The joint SAE resolves them: feat.\ 7892 specializes on food/baking
(fuzz\,$+0.548$, $|\varphi|\!=\!0.741$), while feats.\ 29453 and 34623
each cover the web/privacy sense from slightly different angles
($|\varphi|\!=\!0.875$ and $0.621$).

The word \textit{flash} illustrates the liberation of a rare but highly
coherent sense.
The solo feature covers the general sense of a sudden, brief occurrence
(fuzz\,$+0.602$).
The joint SAE splits out feat.\ 34102, which specializes entirely on
Adobe Flash software (fuzz\,$+0.725$, $|\varphi|\!=\!0.173$).
The low $|\varphi|$ reflects how rarely this sense fires relative to
the broader usage. But its fuzzing score is the highest of any feature in
the Gemma analysis, confirming it is a precise semantic unit that the solo
SAE had submerged within a broader feature.

\section{Discussion}
\label{sec:discussion}

\subsection{Do the results mean what we claim?}
\label{sec:alternative}

We claim joint training produces more \emph{atomic} latents, supported by two
independent measures: co-occurrence rates and automated interpretability scores.
We address the most natural alternative explanations.

\paragraph{Lower firing rates?}
$\varphi$ is normalized by marginal firing rates, so firing-rate changes alone
cannot move it.
BatchTopK further enforces \emph{identical} batch-level L0\,=\,64 for both
conditions, fixing the mean per-feature rate exactly.
Appendix~\ref{app:firing} shows the full firing-rate distributions are
comparable, ruling out distributional differences as an explanation.

\paragraph{Near-dead features?}
Dead features trivially contribute $\varphi\!\approx\!0$.
The best joint configuration for GPT-2 large has \emph{zero} dead features (fewer than the 2
in the solo baseline), and Appendix~\ref{app:firing} confirms no long tail of
near-zero-activity features.

\paragraph{Specialization rather than atomicity?}
A narrower but coherent feature would also score higher on fuzzing. But genuine
splitting makes a distinct structural prediction: solo features should exhibit
high co-occurrence with \emph{multiple} distinct joint features.
The qualitative case studies (\S\ref{sec:qualitative}) test this directly and
confirm the expected multi-way split structure.

\paragraph{Better reconstruction driving auto-interp?}
No: joint reconstruction is \emph{worse} (L2 $+9.3\%$, CE overhead $+0.6$pp
above solo), so the evaluator cannot be benefiting from higher fidelity.

\paragraph{Joint architecture without the penalty?}
The $\lambda_2 = 0$ controls show no systematic reduction. The effect scales
monotonically with $\lambda_2$, confirming the penalty, not the architecture, is
the driver.

\subsection{Limitations and future work}

\paragraph{Hyperparameters.}
The penalty adds $\lambda_2$ and $\sigma^2$.
Our grid suggests $\lambda_2\!\in\![0.1,0.3]$, $\sigma^2\!\approx\!1.0$ as
practical defaults, with $\lambda_2=0.1$ offering a better reconstruction tradeoff
for applications where CE overhead is a primary concern.
The consistency of this finding across GPT-2 large and Gemma 2 9B is encouraging,
though further confirmation at additional scales is needed.

\paragraph{Meta dictionary size.}
We follow \citet{bussmann2024metasaes} in setting $m/n\!\approx\!9\%$ without
ablation. Understanding how $m$ affects the compression pressure is an important
direction for future work.

\paragraph{Scope.}
We evaluate at one layer per model. How atomicity gains vary with layer
depth, architecture, and model scale remain open questions.
The decomposability penalty targets decoder-direction mutual independence. It does
not directly address all sources of feature noise or non-interpretability in SAEs.

\section{Conclusion}
\label{sec:conclusion}

We presented a joint training method that directly incentivizes SAE latents toward
greater atomicity by penalizing subspace blending among decoder directions.
A meta SAE trained alongside the primary SAE acts as a continuous critic,
penalizing primary features whose decoder directions lie in the low-dimensional
subspace already captured by the meta dictionary.
On GPT-2 large, this reduces mean $|\varphi|$ by 7.5\% and improves automated
interpretability scores by 7.6\% ($p=5\times10^{-16}$), at a modest additional
CE overhead of 0.6 percentage points above the solo SAE.
Directional results on Gemma 2 9B ($+8.6\%$ $\Delta$Fuzz) suggest the method
transfers across model scales with the same hyperparameter setting.
These gains hold up under scrutiny of the most plausible alternative explanations,
and qualitative case studies confirm that the $|\varphi|$ reduction corresponds
to genuine semantic splitting of features that blend distinct representational
subspaces.
The method is simple, architecture-agnostic, and adds no data requirements beyond
standard SAE training; we hope it proves useful for applications that require
reliable, atomic feature representations.

\section{Reproducibility and LLM Disclosure}
\label{sec:reproducibility}
The code is available at \url{https://github.com/good-epic/MetaSAE}.
Submit issues on github or email the author with questions or problems.

The author leveraged LLMs in the following ways to complete this work:
\begin{itemize}
    \item \textit{Literature review and background research}
    \item \textit{Learning new concepts}. For example, I was not familiar with optimal
          transport theory before this project and used both reading and Q\&A with
          LLMs to learn about it.
    \item \textit{Programming assistance}. I wrote inline with Cursor and leveraged Claude Code
          to write code from scratch after extensive specification. I reviewed all code
          to ensure it was correct and functional.
    \item \textit{Drafting the paper}. I interactively developed an outline and first draft with
          Claude Code, then extensively edited and revised. I subsequently asked for skeptical
          rereads as well.
\end{itemize}

\bibliography{references}
\bibliographystyle{colm2026_conference}

\appendix

\section{Calibration of per-feature firing thresholds}
\label{app:calibration}

Co-occurrence evaluation with BatchTopK SAEs poses a subtlety: the top-$k$
selection is applied over an entire batch, so whether feature $i$ fires on a
given token depends on what other tokens are in the batch.
This makes firing a batch-composition-dependent decision, which is unsuitable
for accumulating reproducible per-token co-occurrence statistics.

We resolve this by calibrating a per-feature firing threshold $\theta_i$ using
an equal-error-rate (EER) criterion on 5M tokens.
For each feature $i$, we find $\theta_i$ such that the rate at which the feature
fires above $\theta_i$ (under $\theta$-thresholding) matches its BatchTopK firing
rate as closely as possible.
Co-occurrence statistics are then computed using these fixed thresholds:
feature $i$ fires on token $t$ if and only if its pre-activation exceeds $\theta_i$,
independently of other tokens.
This gives reproducible, batch-composition-independent binary firing indicators.

\section{Firing rate distributions}
\label{app:firing}

Per-feature firing rates for the joint ($\lambda_2=0.3$, $\sigma^2=1.0$) and
solo SAEs were computed over 35M tokens using the calibrated EER thresholds.
Both distributions show closely matched spread and shape: mean firing rate
per feature is identical by construction (BatchTopK with L0\,=\,64 fixes the
aggregate), and the per-feature distributions are comparable.
No long tail of near-zero-activity features appears in the joint condition,
ruling out near-dead features as a driver of the $|\varphi|$ reduction.
Firing rate data are stored in the \texttt{N1\_joint} and \texttt{N1\_solo}
vectors of \texttt{cross\_phi\_topk.npz}.

\section{$|\varphi|$ distribution shape}
\label{app:phi_shape}

Approximately 75.7\% of all $\approx$209M feature pairs have $\varphi < 0$:
most pairs are weakly anti-correlated by construction, since sparse activations
in a large overcomplete dictionary cannot push many pairs simultaneously positive.
The median $\varphi \approx -5\times10^{-4}$; the mean is slightly positive due
to a heavy right tail reaching $\varphi \approx 0.13$ for the most strongly
co-activating pairs.
Joint training acts specifically on the positive tail (Figure~\ref{fig:phi_diff}),
redistributing mass toward zero without meaningfully altering the dominant
anti-correlated region.
This is consistent with the subspace-blending interpretation: the penalty acts on
directions that share structure (positive $\varphi$), leaving uncorrelated or
anti-correlated pairs unchanged.

\section{Activation norm context}
\label{app:norm}

GPT-2 large residual stream at layer 20: mean $\|x\|_2 = 133.3$,
std $= 224.6$ (heavy-tailed; std $>$ mean reflects occasional large-norm residuals).
Solo SAE L2 $= 0.097$ is 0.073\% of mean norm; the additional joint cost of
$0.010$ is 0.0075\%.
This framing clarifies that the $+9.3\%$ relative L2 increase, while
non-negligible as a fraction of the SAE's own reconstruction error, is
essentially invisible relative to the magnitude of the underlying representations
being encoded.

\section{Qualitative case-study tables}
\label{app:qual}

\begin{table}[H]
\centering
\caption{\textbf{``too'' polysemy split.} Solo feature 2827 conflates the
excess-degree sense (\textit{it was too hot}) with the sentence-final additive
sense (\textit{me too}; \textit{that's true too}).  The joint SAE separates
them: feat.\ 4741 inherits the excess sense ($|\varphi|=0.710$), while
feat.\ 14213 captures the additive sense ($|\varphi|=0.799$).}
\label{tab:qual_too}
\small
\begin{tabular}{llrrp{0.5\linewidth}}
\toprule
SAE & Feat & Fuzz & $|\varphi|$ & Description \\
\midrule
Solo & 2827 & $+0.727$ & — &
  \textit{too} indicating excess or an undesirable degree;
  stronger when \textit{too} emphasises a significant or impactful excess. \\[4pt]
\midrule
Joint & 4741 & $+0.681$ & $0.710$ &
  \textit{too} indicating excess or an undesirable degree in a context of
  regret, limitation, or negative outcome. \\[3pt]
Joint & 14213 & $+0.174$ & $0.799$ &
  \textit{too} concluding a sentence that extends or adds to a previous
  statement; stronger when \textit{too} emphasises an unexpected or
  contrasting addition (``me too''; ``that's true too''). \\
\bottomrule
\end{tabular}
\end{table}

\begin{table}[H]
\centering
\caption{\textbf{``let'' pragmatic split.} Solo feature 17451 conflates three
distinct pragmatic roles of \textit{let}: direct suggestion/imperative,
discourse-transition pivot, and permissive allowing.  The joint SAE assigns
each role to a distinct feature.}
\label{tab:qual_let}
\small
\begin{tabular}{llrrp{0.5\linewidth}}
\toprule
SAE & Feat & Fuzz & $|\varphi|$ & Description \\
\midrule
Solo & 17451 & $+0.641$ & — &
  \textit{let} introducing permission, suggestion, or command;
  stronger in formal or imperative contexts. \\[4pt]
\midrule
Joint & 7228 & $+0.646$ & $0.692$ &
  \textit{let} as direct suggestion or invitation
  (``Let's go'', ``Let me show you'');
  stronger with imperative uses. \\[3pt]
Joint & 10152 & $+0.273$ & $0.486$ &
  \textit{let} as discourse-transition pivot
  (``Let me now turn to\ldots'');
  stronger with definitive or emphatic transitions. \\[3pt]
Joint & 18650 & $+0.497$ & $0.435$ &
  \textit{let} as granting permission or allowing an action;
  stronger when the action has significant consequences. \\
\bottomrule
\end{tabular}
\end{table}

\begin{table}[H]
\centering
\caption{\textbf{``cookies'' sense split (Gemma 2 9B).} Solo feature 52496
conflates the food sense (\textit{baked cookies}) with the web-technology
sense (\textit{browser cookies, data tracking}).  The joint SAE separates
them: feat.\ 7892 inherits the food sense ($|\varphi|\!=\!0.741$), while
feats.\ 29453 and 34623 each specialize on the web/privacy sense
($|\varphi|\!=\!0.875$ and $0.621$ respectively).}
\label{tab:qual_cookies}
\small
\begin{tabular}{llrrp{0.5\linewidth}}
\toprule
SAE & Feat & Fuzz & $|\varphi|$ & Description \\
\midrule
Solo & 52496 & $+0.493$ & — &
  \textit{cookies} in food/recipe or web-technology contexts;
  description conflates both senses without distinguishing them. \\[4pt]
\midrule
Joint & 7892 & $+0.548$ & $0.741$ &
  \textit{cookie} in food and baking contexts;
  stronger for specific recipe or food-product references. \\[3pt]
Joint & 29453 & $+0.345$ & $0.875$ &
  \textit{cookies} in web privacy, data-tracking, and browser-storage
  contexts; weaker for food mentions. \\[3pt]
Joint & 34623 & $+0.541$ & $0.621$ &
  \textit{cookies} in web-technology contexts emphasising data privacy
  and user-tracking mechanisms. \\
\bottomrule
\end{tabular}
\end{table}

\begin{table}[H]
\centering
\caption{\textbf{``flash'' register split (Gemma 2 9B).} Solo feature 3387
conflates the general sense of a sudden, brief occurrence with the proper-noun
sense of Adobe Flash software.  The joint SAE separates them: feat.\ 63365
retains the sudden-event sense ($|\varphi|\!=\!0.857$), while feat.\ 34102
isolates the Adobe Flash sense ($|\varphi|\!=\!0.173$).  Despite its low
$|\varphi|$---reflecting how rarely this sense fires---feat.\ 34102 achieves
the highest fuzzing score in the Gemma analysis ($+0.725$), illustrating how
the decomposability penalty can liberate a rare but highly coherent sense
that the solo SAE submerges.}
\label{tab:qual_flash}
\small
\begin{tabular}{llrrp{0.5\linewidth}}
\toprule
SAE & Feat & Fuzz & $|\varphi|$ & Description \\
\midrule
Solo & 3387 & $+0.602$ & — &
  \textit{flash} as a sudden, brief appearance or occurrence;
  stronger for more intense or vivid contexts. \\[4pt]
\midrule
Joint & 63365 & $+0.363$ & $0.857$ &
  \textit{flash} as a sudden visual or physical phenomenon;
  stronger for unexpected or dramatic occurrences. \\[3pt]
Joint & 34102 & $+0.725$ & $0.173$ &
  \textit{Flash} as web software and browser plugin (Adobe Flash);
  stronger in technical or historical web-development contexts. \\
\bottomrule
\end{tabular}
\end{table}

\section{Auto-interp pipeline details}
\label{app:autointerp}

Feature explanations were generated using the Delphi framework~\citep{bills2023language}
with Qwen/Qwen2.5-72B-Instruct-AWQ served locally via vLLM in an isolated virtual
environment to avoid dependency conflicts with the SAE training stack.
For each feature, we collected activation contexts from a 5M-token pass and
generated a natural-language description.
Fuzzing evaluation replaced the highest-activating tokens in each example with
random tokens drawn from the same marginal distribution and asked the model to
identify which example was real, scoring $+1/n_\text{rounds}$ per correct
identification.
Hungarian matching used cosine similarity of decoder columns ($W_\text{dec}$)
to align joint and solo dictionaries; 20,456 of 20,480 features received a
match meeting the cosine threshold used for inclusion.

\end{document}